# Physical Signal Classification Via Deep Neural Networks

Benjamin Epstein[1], PhD, Roy H. Olsson III[2], PhD

[1]ECS Federal, Arlington, VA
[2]Defense Advanced Research Projects Agency (DARPA), Arlington, VA

*Abstract*—A Deep Neural Network is applied to classify physical signatures obtained from physical sensor measurements of running gasoline and diesel-powered vehicles and other devices. The classification provides information on the target identities as to vehicle type and even vehicle model. The physical measurements include acoustic, acceleration (vibration), geophonic, and magnetic.

## Introduction

Deep Neural Networks (DNNs) have a broad variety of applications where time stepped sensor data can be used to train and classify target types. We support this claim by training a DNN to identify and classify automotive vehicle types through the use of physical data obtained from accelerometer, acoustic, geophone, and magnetic measurements of the vehicles at various distances and orientations from the vehicles. Preprocessing of the measurement data, followed by a three-layer DNN implemented in TensorFlow, yielded promising results. Our effort also shows that the choice of a pre-processing strategy is critical to obtaining highly correct classification rates. This work follows earlier work in sensor data classification [1,2,3], but with newer tools and techniques.

The overall classification scheme for the DNN determination of the target type is provided in Figure 1. The various steps of Figure 1 are described in the text to follow.

.

## Methods

A. *Physical Measurements*

Physical measurements from a variety of automotive types were collected by MIT Lincoln Labs as part of an effort under the DARPA Near Zero RF and Sensor Operations (N-ZERO) program [4]. N-ZERO aims to develop wake-up circuits for unattended ground sensor devices that consume at most only 10nW of power while "listening" for the presence of signals of a characteristic wake-up profile. N-ZERO includes sensors that watch for either RF or physical signals of a given target type. The physical measurements were obtained for the study of the discrimination performance of physical signal sensors developed during the N-ZERO program. Interests in applying neural networks as part of the target classification scheme prompted the effort reported herein to better understand the applicability of DNNs to assist classification of the sensor response.

The targets of the effort are listed in Table 1. These targets operated in an idle state during the measurements, with one target at a time running. Such targets were deemed representative of the vehicle types to be encountered by the N-ZERO unattended ground sensors. Table 2 summarizes the types of physical measurements and equipment applied. Measurements were conducted in both quiet conditions (inside an anechoic chamber) and outdoors under realistic ambient environments.

B. *Data Pre-pro*cessing

Five to ten data files containing the physical measurements from each target type were randomly selected as sources of training and test/validation data. Each file contained several million sample points in time, acquired at 20 µsec sampling intervals corresponding to a 50k samples/sec sampling rate. Each time point contained data from all 13 physical sample measurements (as described in Table 2). Twenty, roughly one-sec data blocks were randomly extracted from each data file, with a Fast Fourier Transform (FFT) computed over each of the extracted 1-sec blocks. Since no target-identifying frequency domain signature could be ascertained for frequencies in the FFT above 300 Hz, frequency domain data from only 300 Hz and below were preserved for all physical measurement types, with the remaining data above 300 Hz removed from the analysis.

Using the remaining 300 lower frequency spectral bins, heat maps were generated showing the spectral response for each sensor type over an average of about 100 randomly selected 1-sec intervals for each target type (see Figure 2). By visually comparing the heat maps for the different target types, it becomes apparent which physical sensor data have more orthogonality in identifying the targets. Through such visual inspection, it was determined that only some of the physical measurements appeared to correspond to unique target signatures; these are indicated in Table 3. Of course, a less subjective numerical approach could replace the visual inspection by mathematical means, such as a principle components analysis.

Using a Matlab script, the FFT spectra were computed from typically 1000 randomly collected 1-sec time intervals of sensor collections, where the FFTs now were applied only to the useful physical sensor data drawn from the sensor measurements of Table 3. As the FFTs of each sensor were computed, the resulting frequency points for each sensor type were fused by computing their weighted average:

$$FreqPoint_i = \sum_{j=1}^{N} w_j \cdot |S_{ij}| \quad (eq. 1)$$

where the ith frequency point $FreqPoint_i$ is computed as a weighted average of the N sensor response FFT values $S_{ij}$, where $i$ corresponds to the frequency (a total of 300 frequency points were computed for each 1-sec interval)) and $j$ corresponds to each of the N selected sensor types (e.g., accelerometer, geophone, etc.) at that frequency point. Admittedly, different combinations of the sensor measurements, or keeping the sensor data types separate, likely will show different frequency-dependent profiles which could be exploited to achieve better DNN accuracy. However, the rather ad hoc approach taken was pursued mainly because it simplified the later steps of the analysis. Separating the measurement types and fusing the DNN results should be the subject of later study.

The next step determined which fused spectral components added information to the target classification. Again, a visual inspection of the target heat maps might help to answer this; however a more methodical approach was applied:
1) Compute mean response, for each target type, at each of the 300 frequency points, across hundreds of rows (i.e., 1-sec interval FFTs).
2) Compute mean response over all target types collectively at each of the 300 frequency points.
3) Compute ratio of mean for each target type over collective mean over all target types, at each frequency.
4) If the above ratio exceeds a threshold, include this frequency point in the DNN training and testing. A threshold value of 1.5 to 2 was found to work well.
5) In the neural net of this study, 20 to 125 spectral points (of the 300 saved) were selected in the above manner, depending on the number of vehicle types to be classified.

In determining a threshold value for Step 4 above, we need to select a distribution of over-threshold values that are spread over the frequencies and target types in such a manner as to uniquely identify the target types. In other words, we do not want any frequency bin to have too many super-threshold values down its column since such frequencies would not be unique to a given target type.

C. *Apply DNN*

The DNN was implemented in Python using various Python mathematical support libraries and TensorFlow, drawing from methodologies described in [5]. The architecture calls for a three layer structure with an input layer assigning one neuron to each frequency component $FreqPoint_i$ (eq. 1) after the reduction to useful frequency components, as described above (resulting in 20 to 125 frequency points). The DNN also contains a hidden layer with the same number of neurons as the input layer, and output layer whose number of neurons and outputs correspond to the target types to be identified. The three DNN layers were constructed such that inputs and outputs of each layer were determined by the matrix relationship:

$$[Out] = [Weight][X] + [Bias] \quad (eq. 2)$$

where the column vector *[Out]* of dimension n is the product of an n x m matrix *[Weight]* and input value column vector *[X]* of dimension m, with the addition of a bias term *[Bias]* column vector of dimension n. The first two layers make use of a sigmoid threshold function on their outputs, while the third (output) layer applies the direct matrix operation output. For the results reported below, the Group 1 classification applied about 115 frequency points as the feature set inputs to the DNN; Group 2 used 23 frequency point inputs. The feature set size could be flexible depending on the number of training iterations, tweaks in the weighting of the sensor measurements, and other factors. Remember, each of the first two layers contained the same number of nodes, with the third (output) layer corresponding to the number of categories in the classification.

During training, the goal is to determine the values of the *[Weight]* and *[Bias]* vectors that best matches the training target types to the predicted target types given the training data feature sets (i.e., fused physical sensor frequency components). This match is achieved through a use of the "Adam Optimizer" [6], which is a type of stochastic gradient descent optimizer (where the TensorFlow *AdamOptimizer* settings were at default values with learn rates of 0.005). The objective function applied in the optimization implements the TensorFlow *tf.nn.sigmoid_cross_entropy_with_logits* function. In effect, this function computes the probability of error in the comparison of the computed vs. given target values across the randomly selected training set data. Once this value is computed for each training set (row), the mean cross entropy across all of the applied training sets serves as the loss value that is minimized (and plotted in Figure 3).

The Python / TensorFlow program executed the following steps:

Step 1   Read in all ~1000 data rows. All 300 frequency points are read in for each row (corresponding to the spectra of the randomly selected one second intervals); however, only those frequency points selected in the methodology described above are applied in the analysis.
Step 2   Convert target identifiers, stored in the $301^{st}$ column of the input data, to one-hot encoding.
Step 3   Randomly select 80% of the rows for training, with the remaining rows for testing. This results in ~800 rows of data available for training; ~200 rows for testing. Random selection is performed without substitution, thereby avoiding any overlap between training and test sets.
Step 4   Of the 800 training rows, randomly select a "batch" of 150 rows.
Step 5   Perform backpropagation over the 150 rows to arrive at neural net weights and biases that best match the computed target to the actual target. Compute the loss function which compares actual target one-hot values against computed one-hot target values.
Step 6   Using the computed weights, apply them to the 200 test rows. Compare computed test output (prediction) against actual test output.
Step 7   Repeat the above process a given number of times – typically about 200 to 1000 times. Each cycle of applying 150 training points to determine the weights and bias values, and applying those results to the 200 test points, constitutes a "run". For each run select a different set of 150 rows from the 800 rows available for training. Store computed loss between actual and computed results for training and test round. Plot these loss values.
Step 8   Compute "accuracy". This is a tally of the matches in the one-hot computed and actual target vales (applied to training and test data).
Step 9   Compute confusion matrix.

**Results**

Figure 3 shows the convergence of the accuracy from representative DNN runs for the Group 1 (vehicle type) and Group 2 (all quiet, generator, small car, truck) tests. It is clear that after about 200 training runs, not much improvement in accuracy is gained and that the DNN becomes "over trained" on the training sets. Of course, the convergence changes slightly in one set of runs against another due to the use of different training rows, but the fact that all convergences are similar points to some level of robustness of the training method. Figure 4 provides a confusion matrix describing the classification of the different target types for the Group 1 and Group 2 runs. About 30 of 200 test rows (15%) are incorrectly classified (Group 1), and only 6 of 128 test runs for Group 2 are misclassified.

**Discussion**

A simple three-level DNN shows promise in its ability to classify the physical signatures of different automotive types. This result should be useful for identifying vehicle types by unattended ground sensors, such as those located at security check points or along roads to assess traffic conditions and road usage. We emphasize that the results presented herein are only preliminary, and do not account for the myriads of ways the sensor data can be pre-analyzed, not to mention how the DNN can be constructed, trained and tested. Nevertheless, this preliminary effort points to many possibilities on how sensor data can be fused and reduced as a means of simplifying the unattended ground sensor DNN schemes.

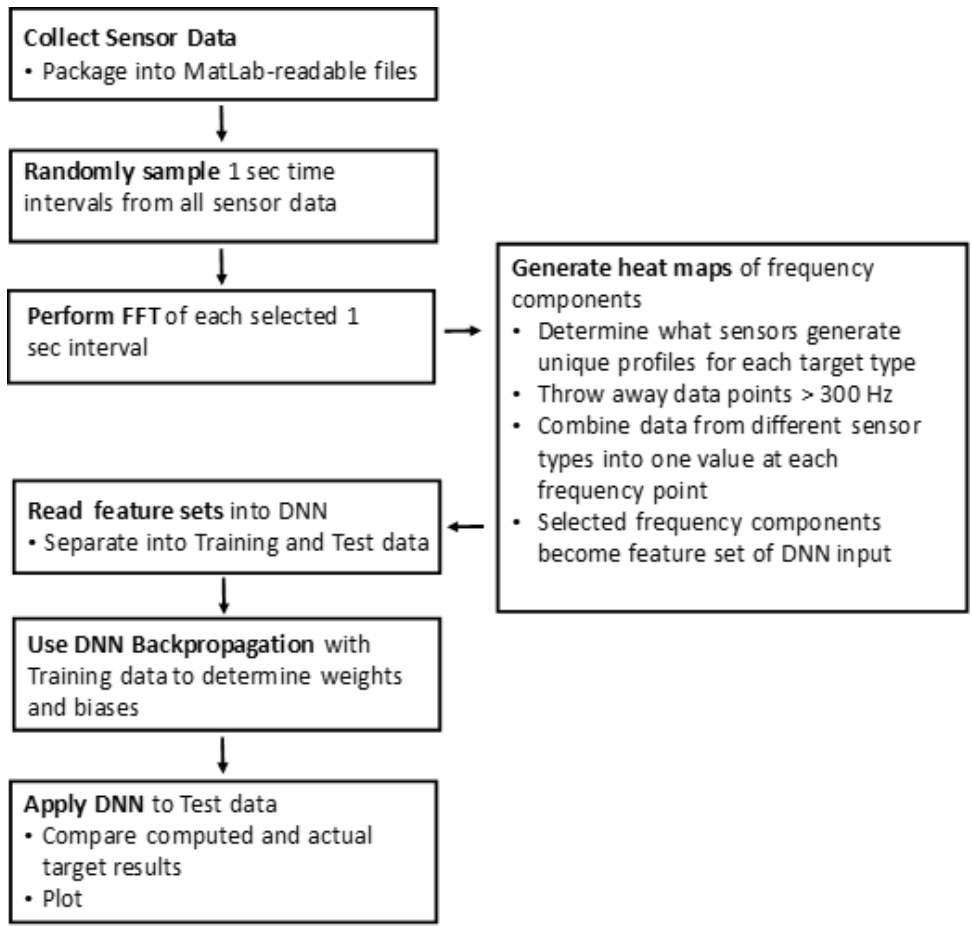

**Figure 1.** Overall flow of target classification process.

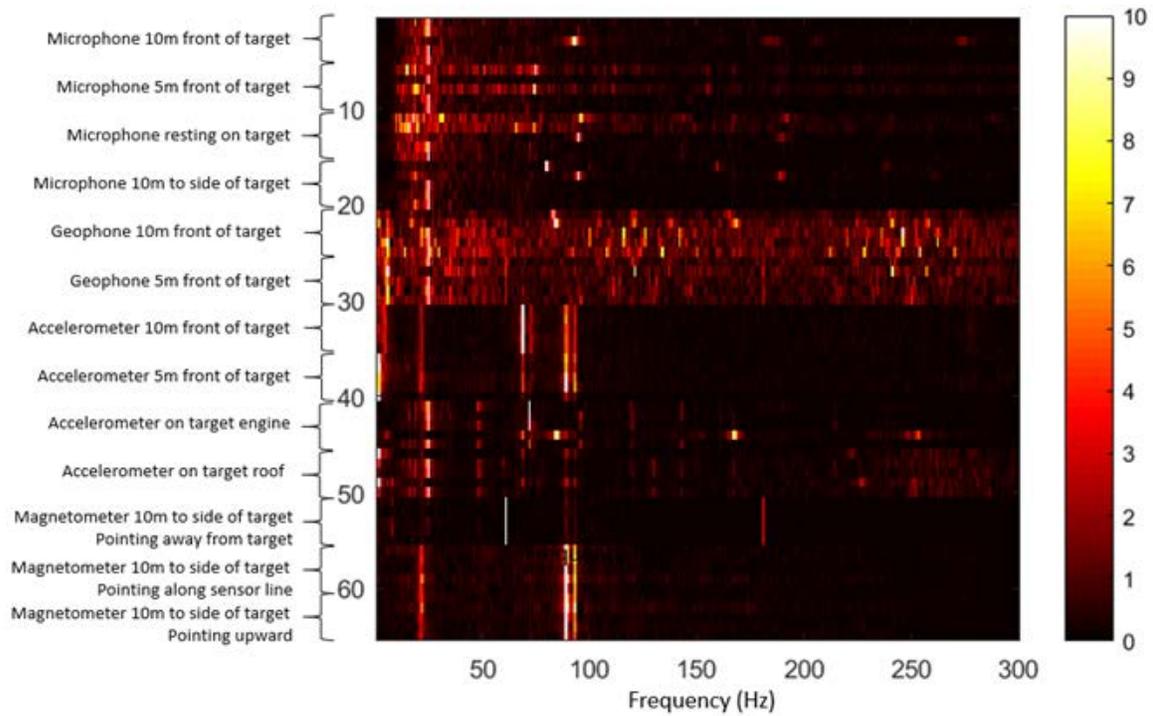

**Figure 2.** Heat map of one target type (Honda Civic in rural setting). The vertical axis corresponds to the types of sensor measurements, as listed in Table 2; all values were normalized to the peak value of the measurement on a scale of 0 to 10 (hence, the values are unit-less). The horizontal axis is computed FFT frequency (in Hz). There are typically 5 sets of measurements for each sensor type and position, hence multiple rows for each sensor type.

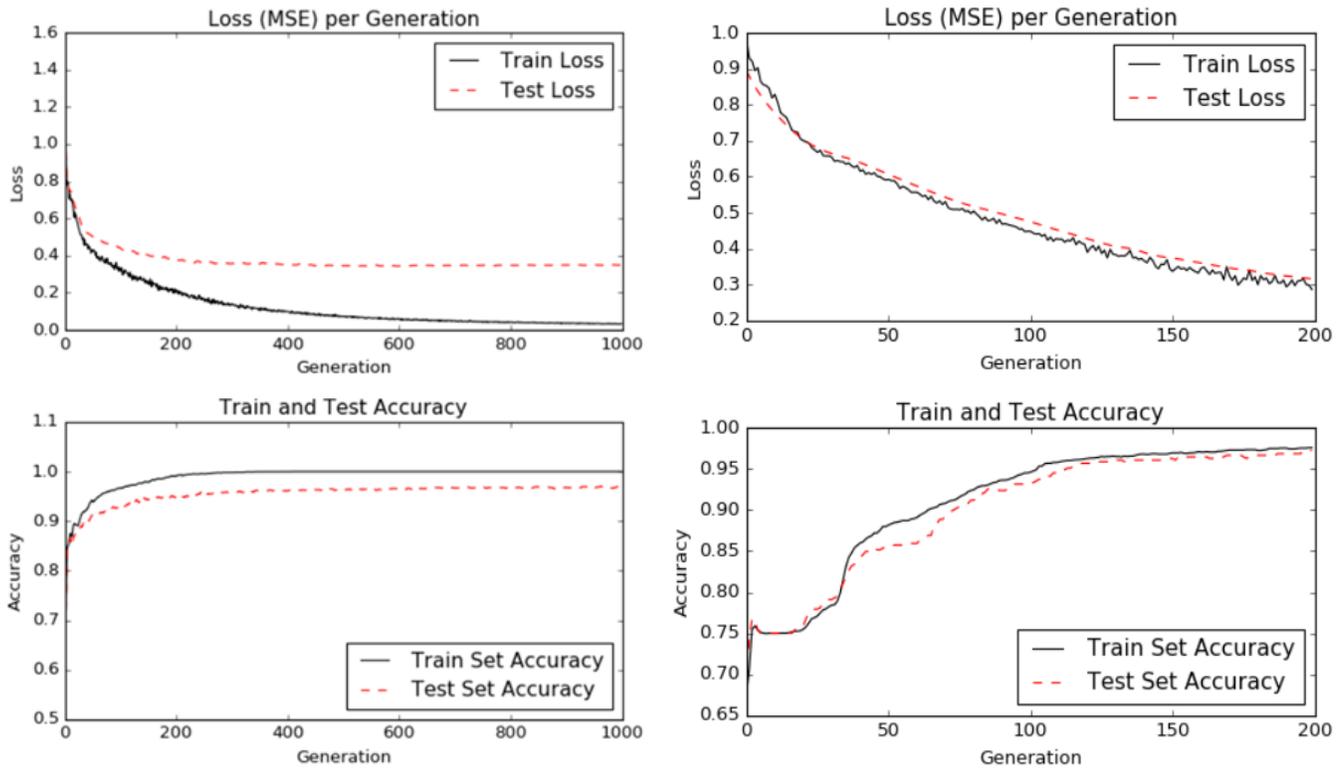

**Figure 3.** Progress of training runs (left two graphs) for Group 1 target classifications, (right two graphs) Group 2 target classifications. Loss plot shows progress of loss function. Training test and accuracy compares agreement of 1s and 0s in one-hot representations of target and computed classification data.

|  | All Quiet | Honda Civic | Toyota Corolla | Ford F-150 Pick | Diesel | Ford Fusion | Acura MDX | Unclassified |
|---|---|---|---|---|---|---|---|---|
| All Quiet | 32 | 0 | 2 | 0 | 0 | 0 | 0 | 8 |
| Honda Civic | 0 | 21 | 0 | 0 | 0 | 0 | 1 | 1 |
| Toyota Corolla | 1 | 0 | 15 | 0 | 1 | 1 | 0 | 4 |
| Ford F-150 Pickup Truck | 0 | 0 | 0 | 39 | 0 | 0 | 0 | 2 |
| Diesel | 1 | 0 | 0 | 0 | 11 | 1 | 0 | 0 |
| Ford Fusion | 0 | 0 | 0 | 0 | 0 | 36 | 0 | 0 |
| Acura MDX | 0 | 0 | 0 | 0 | 0 | 0 | 18 | 5 |

|  | All Quiet | Honda Generator | Ford F-150 Pickup Truck | SAAB '83 Car | Unclassified |
|---|---|---|---|---|---|
| All Quiet | 26 | 0 | 0 | 1 | 5 |
| Honda Generator | 0 | 33 | 0 | 0 | 0 |
| Ford F-150 Pickup Truck | 0 | 0 | 30 | 1 | 3 |
| SAAB '83 Car | 0 | 0 | 0 | 28 | 1 |

**Figure 4.** Confusion matrices for Group 1 (vehicle type tests) (top) and Group 2 (all quiet, generator, small car, truck) (bottom) tests.

**TABLE 1. TARGET TYPE DESCRIPTION**

    **Group 1:**
        All Quiet
        Honda Civic
        Toyota Corolla
        Ford F150 Pickup Truck
        Mercedes Diesel "Sprinter" Van
        Ford Fusion
        Acura MDX
    **Group 2:**
        All Quiet
        Honda Generator
        Ford F-150 Truck
        SAAB 83 Car

**TABLE 2. SENSOR MEASUREMENTS**

Microphone[1] 10 meters to the front of target (in pascals)
Microphone 5 meters to the front of target (in pascals)
Microphone on the target (in pascals)
Microphone 10 meters to the side of the target (in pascals)
Geophone[2] 10 meters to the front of target (in m/s)
Geophone 5 meters to the front of target (in m/s)
Accelerometer[3] 10 meters to the front of target (in m/s^2)
Accelerometer 5 meters to the front of target (in m/s^2)
Accelerometer on the target engine (in m/s^2)
Accelerometer on the target roof in (m/s^2)
Magnetometer[4] x axis, 10 meters to the side of target, pointing away from target (in tesla)
Magnetometer y axis, 10 meters to the side of target, pointing along sensor line (in tesla)
Magnetometer z axis, 10 meters to the side of target, pointing up (in tesla)

---

[1]Earthworks M23
[2]Geospace device
[3]PCB 356B18
[4]Applied Physics Systems
NOTE: measurements where sensors were physically placed on the targets were not used in the analysis.

**TABLE 3. SENSOR MEASUREMENTS SELECTED FOR ANALYSIS**
    **Group 1:**
Microphone 10 meters to the front of target (pascals)
Microphone 10 meters to the side of the target (pascals)
Geophone 10 meters to the front of target (m/s)
Accelerometer 10 meters to the front of target (m/s^2)
    **Group 2:**
Geophone 10 meters to the front of target (m/s)
Accelerometer 5 meters to the front of target (m/s^2)
Magnetometer z axis, 10 meters to the side of target, pointing up (tesla)

**AUTHOR BIOGRAPHIES**

**Benjamin R. Epstein** received the B.S. degree in electrical engineering from the University of Rochester, Rochester, NY, USA, in 1978, the Ph.D. degree from the School of Engineering and Applied Sciences, University of Pennsylvania, Philadelphia, PA, USA, in 1983, and an MBA from Stern School of Business, New York University, New York, NY, USA, in 1991. He is currently a Senior Scientist at ECS Federal, Arlington, VA, USA and serves as a Senior Advisor to the Defense Advanced Research Projects Agency (DARPA) Microsystems Technology Office in subjects pertaining to phased arrays, low power sensor electronics, low frequency transmission, microcircuit security, and other areas. As VP Special Projects of OpCoast LLC, his areas have spanned electronic warfare and electronic–biological interfaces, including insect-based ad hoc communications networks. He served as VP of Business Development at the U.S. operation of Orange (formerly France Telecom), overseeing their investment activities in North America and the rollout of new international satellite and fiber-based data communications and video services. He also spent eight years as Member Technical Staff at Sarnoff Corp. (formerly RCA Laboratories), where he executed projects involving RF/microwave systems design, smart skin array antennas, simulation, and testing, as well as supercomputer applications in imaging analysis. He also served as a Postdoctoral Fellow at the Centre National de la Recherche Scientifique (CNRS) in France. He has numerous publications on the topics of nonlinear simulation, circuit fault analysis, biomachine interfaces, and RF/microwave propagation. Dr. Epstein is a member of Sigma Xi, Tau Beta Pi, and the American Association for the Advancement of Science (AAAS). His IEEE activities incudes the role of Publications Chairman for past and current IEEE International Conference on Microwaves, Communications, Antennas, and Electronic Systems (COMCAS).

**Roy H. Olsson III** received B.S. degrees (Summa Cum Laude) in electrical engineering and in computer engineering from West Virginia University in 1999 and the MS and Ph.D. degrees in electrical engineering from the University of Michigan, Ann Arbor in 2001 and 2004. He is currently a program manager in the Microsystems Technology Office (MTO) at the Defense Advanced Research Projects Agency (DARPA). His research interests include materials, devices, and architectures for low-power processing of wireless and sensor signals, miniature antennas, and phased array antennas. Prior to joining DARPA, he was a Principal Electronics Engineer in the MEMS Technologies Department at Sandia National Laboratories in Albuquerque, NM. At Sandia, he led research programs in aluminum nitride and lithium niobate piezoelectric micro-devices for processing of RF, inertial and optical signals. He has authored more than 100 technical journal and conference papers and holds 27 patents in the area of microelectronics and microelectromechanical systems (MEMS). He served on the organizing committee of the 2011 Phononics Conference and was a Member of the Technical Program Committee for the IEEE Ultrasonics Symposium (IUS) from 2010-2016. He is a member of the IEEE Solid State Circuits Society; the IEEE Ultrasonics, Ferroelectrics, and Frequency Control Society; Eta Kappa Nu; and Tau Beta Pi. He was awarded an R&D100 award in 2011 for his work on Microresonator Filters and Frequency References and was named the 2017 DARPA program manager of the year.

**ADDITIONAL INFORMATION**

**Competing interests**

The authors declare no competing interests.

**Author Contributions**

Dr. Benjamin Epstein, the Corresponding Author for this submission, developed and performed the neural network analysis described herein, and wrote the initial draft of the manuscript. Dr. Roy Olsson provided editorial insight, corrections, and was instrumental in formulating the data acquisition.

This manuscript has been approved for public release by the US Defense Advanced Research Projects Agency (DARPA).